\documentclass[10pt,twocolumn,letterpaper]{article}

\usepackage{cvpr}              % To produce the CAMERA-READY version
\usepackage{times}
\usepackage{epsfig}
\usepackage{graphicx}
\usepackage{amsmath}
\usepackage{amssymb}
\usepackage{booktabs}
\usepackage{multirow}
\usepackage{caption}
\usepackage{subcaption}
\usepackage{marvosym}
\usepackage[accsupp]{axessibility} % Improves PDF readability for those with visual impairments.

\usepackage[dvipsnames]{xcolor}

\definecolor{cvprblue}{rgb}{0.21,0.49,0.74}
\usepackage[pagebackref,breaklinks,colorlinks,citecolor=cvprblue]{hyperref}

\def\paperID{11775} % *** Enter the Paper ID here
\def\confName{CVPR}
\def\confYear{2024}

\title{PTM-VQA: Efficient Video Quality Assessment Leveraging \\ Diverse PreTrained Models from the Wild}

\author{Kun Yuan$^{1\dag}$, ~Hongbo Liu$^{1,2\dag}$, ~Mading Li$^{1\dag}$, ~Muyi Sun$^{3,4}$, \\
~Ming Sun$^{1(\textrm{\Letter})}$, ~Jiachao Gong$^{1}$, ~Jinhua Hao$^{1}$, ~Chao Zhou$^{1}$, ~Yansong Tang$^{2(\textrm{\Letter})}$ \\
\textsuperscript{\rm 1}Kuaishou Technology,~
\textsuperscript{\rm 2}Tsinghua University,~
\textsuperscript{\rm 3}School of AI, BUPT,~
\textsuperscript{\rm 4}CASIA \\
{\tt \footnotesize \{yuankun03,limading,sunming03\}@kuaishou.com, liuhbleon@gmail.com, tang.yansong@sz.tsinghua.edu.cn}}

\begin{document}
\maketitle

\renewcommand{\thefootnote}{}
\footnotetext{$^{\dag}$ Equal contribution. $^{\textrm{\Letter}}$ Corresponding authors.}

\begin{abstract}

Video quality assessment (VQA) is a challenging problem due to the numerous factors that can affect the perceptual quality of a video, \eg, content attractiveness, distortion type, motion pattern, and level. However, annotating the Mean opinion score (MOS) for videos is expensive and time-consuming, which limits the scale of VQA datasets, and poses a significant obstacle for deep learning-based methods. In this paper, we propose a VQA method named PTM-VQA, which leverages PreTrained Models to transfer knowledge from models pretrained on various pre-tasks, enabling benefits for VQA from different aspects.

Specifically, we extract features of videos from different pretrained models with frozen weights and integrate them to generate representation. Since these models possess various fields of knowledge and are often trained with labels irrelevant to quality, we propose an Intra-Consistency and Inter-Divisibility (ICID) loss to impose constraints on features extracted by multiple pretrained models. The intra-consistency constraint ensures that features extracted by different pretrained models are in the same unified quality-aware latent space, while the inter-divisibility introduces pseudo clusters based on the annotation of samples and tries to separate features of samples from different clusters. Furthermore, with a constantly growing number of pretrained models, it is crucial to determine which models to use and how to use them. To address this problem, we propose an efficient scheme to select suitable candidates. Models with better clustering performance on VQA datasets are chosen to be our candidates. Extensive experiments demonstrate the effectiveness of the proposed method. 

\end{abstract}

\section{Introduction}

In recent years, social network platforms that focus on videos have gained immense popularity. According to Cisco's Visual Networking Index (VNI), global IP video traffic is predicted to account for 82\% of all IP traffic by 2022, both in business and consumer sectors \cite{barnett2018cisco}. The significant increase in video content consumption poses significant challenges for video providers to deliver better services. Since the perceptual quality of videos has a significant impact on the Quality of Experience (QoE), identifying the quality of videos has become one of the most important issues \cite{DBLP:conf/softcom/KlinkU20,DBLP:journals/tbc/ChikkerurSRK11,DBLP:journals/ejivp/ShahidRLZ14,DBLP:journals/mta/FanLXLH19,DBLP:journals/comsur/ChenWZ15}.
Video quality assessment (VQA) aims to assess the perceptual quality of input videos automatically, imitating the subjective feedback of humans when viewing a video. It has been extensively studied in the context of assessing compression artifacts, transmission errors, and overall quality \cite{DBLP:journals/tip/SaadBC14,DBLP:conf/mm/LiuDW18,DBLP:journals/tip/MittalSB16,DBLP:journals/tip/Korhonen19,DBLP:conf/mm/LiJJ19}. Data-driven deep learning-based methods have been drawing more and more attention compared to conventional methods based on hand-crafted features, as they possess better performance \cite{DBLP:journals/tip/ChenLWDS22,DBLP:conf/mm/ChenLMWS20,DBLP:conf/mm/XuLZZW021,DBLP:journals/access/KossiCDG22,DBLP:journals/tcsv/ChenZLLFW22,DBLP:conf/mm/You21,DBLP:journals/ieeemm/QianPZZLY021,DBLP:journals/ijcv/LiJJ21,DBLP:conf/cvpr/WangKTYBAMY21}.

\begin{figure}[t]
    \centering
    \includegraphics[width=\linewidth]{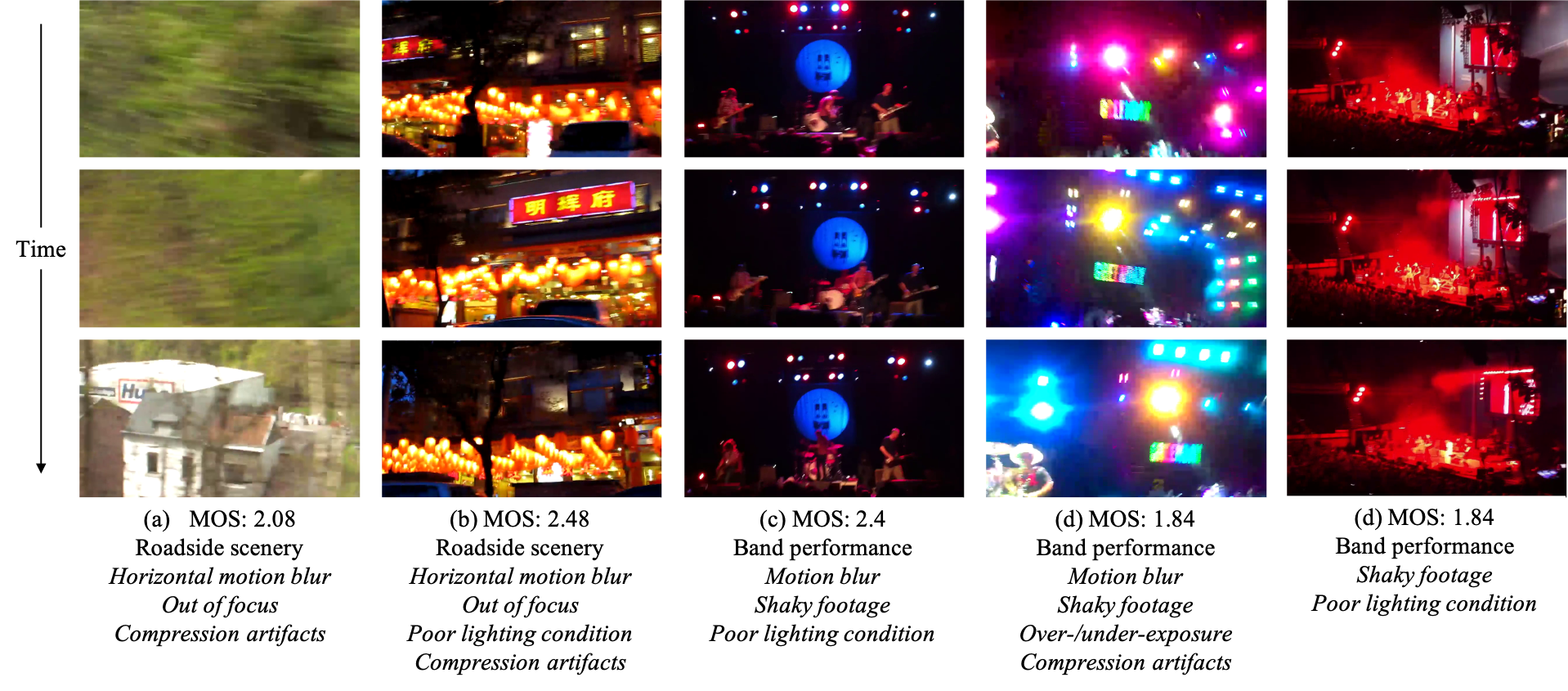}
    \caption{The KoNViD-1k dataset provides video frames that demonstrate a correlation between content/motion patterns and video quality. To identify potential reasons for poor perceptual video quality, we have highlighted specific factors in italics that correspond to the labeled MOS.}
    \label{fig:example}
    \vspace{-0.4cm}
\end{figure}

Compared with other high-level computer vision tasks, datasets for VQA are much smaller. One of the most popular datasets for human action classification Kinetics \cite{DBLP:journals/corr/abs-1907-06987} has 650,000 clips, while the popular VQA dataset KoNViD-1k \cite{DBLP:conf/qomex/HosuHJLMSLS17} has only 1,200 videos. One of the reasons is because VQA is a highly subjective task \cite{DBLP:journals/sigpro/Winkler99,wang2007video}. 
% \textbf{citation}
To obtain an unbiased label, it is recommended by annotation guidelines \cite{rec2006p} that the subjective quality of a single video should be measured in a laboratory test by calculating the arithmetic mean value of multiple subjective judgments, \ie, Mean Opinion Score (MOS). \textit{Take KoNViD-1k as an example, it has 114 votes for each video on average.} This significantly raises the cost of labeling and limits the size of the VQA dataset.
Such a small amount of data limits the power of data-driven VQA methods. To deal with the problem, most existing VQA methods \cite{DBLP:conf/mm/You21,DBLP:journals/tcsv/ChenZLLFW22,DBLP:journals/access/KossiCDG22,DBLP:conf/mm/XuLZZW021} choose to finetune using weights pretrained on common larger datasets (\eg, ImageNet \cite{DBLP:conf/cvpr/DengDSLL009}). 
However, existing works \cite{DBLP:journals/tcsv/LiZTZW22,DBLP:conf/mm/LiJJ19,DBLP:conf/cvpr/WangKTYBAMY21} show that the perceptual quality of a video is related to many factors, \eg, content attractiveness, aesthetic quality, distortion type, motion pattern, and level. Only considering content-based pretrained models may not be sufficient for VQA. Thus, in this work, we focus on how to better utilize a large amount of available pretrained models to benefit VQA.

The present study initially observes a correlation between VQA tasks and other computer vision tasks. To illustrate, Fig.~\ref{fig:example} displays several examples from the KoNViD-1k dataset. It is reasonable to assume that models pretrained on datasets for various pre-tasks have the ability to capture distinct characteristics regarding video quality. 
Consequently, we conduct a simple clustering experiment utilizing Large Margin Nearest Neighbor (LMNN) \cite{DBLP:journals/jmlr/WeinbergerS09} to investigate the correlation between typical pretrained models and the VQA task. Based on the findings, we propose a practical approach, named \textit{PTM-VQA (PreTrained Models-VQA)}, which leverages pretrained models as feature extractors and predicts the quality of input videos based on integrated features. As the parameters of pretrained models remain fixed, we can introduce more pretrained models without exhausting computational resources.

Moreover, we notice that labels in common datasets for pretraining are quite quality-irrelevant. For instance, \textbf{a clear photo of a puppy} with high quality and \textbf{a blurred photo of a puppy} may have the same object-wise label, whereas their quality-wise label may be significantly different. This will confuse the learning process for the VQA task. Therefore, we propose \textit{an Intra-Consistency and Inter-Divisibility (ICID) loss}, which applies constraints on features extracted by multiple pretrained models from different samples. Specifically, model-wise intra-consistency requires features extracted by different pretrained models to be in the same unified quality-aware latent space. Meanwhile, sample-wise inter-divisibility introduces pseudo clusters based on the MOS of samples and aims to separate features of samples from different clusters. 

Furthermore, as the number of pretrained models continues to grow (e.g., PyTorch image models library (Timm) \cite{rw2019timm} supports over 700 pretrained models), finding models suitable for the VQA task through trial-and-error becomes unfeasible. Therefore, we propose to use the Davies-Bouldin Index (DBI) \cite{DBLP:journals/pami/DaviesB79} to evaluate the clustering results and adopt it as the basis for model selection and weighting for feature integration.
To summarize, the \textbf{main contributions} are specified below:
\begin{itemize}
    \item % To the best of our knowledge, this is the first study of how to benefit VQA tasks from a lot of models pretrained on datasets of various tasks. 
    % We explore and verify the correlation between models pretrained on different pre-text tasks and the VQA task and propose a practical NR-VQA method, named PTM-VQA, to exploit cutting-edge pretrained models with diversity to benefit VQA effectively.
    We explore and confirm the association between pretrained models utilizing various pre-text tasks and their effectiveness in performing VQA. Moreover, we present a practical non-reference VQA method named PTM-VQA, which exploits cutting-edge pretrained models with diversity to benefit VQA effectively.
    \item To constrain features with diversity into a unified quality-aware space and eliminate the mismatch between objective and perceptual annotations, we propose an ICID loss. To avoid looking for a needle in a haystack, we propose an effective way to select candidate models based on DBI, which also determines the contributions of different pretrained models.
    % DBI also can be used for integrating features extracted by different pretrained models.
    \item PTM-VQA achieves SOTA performance with a rather small amount of learnable weights on three NR-VQA datasets, including KoNViD-1k, Live-VQC, and YouTube-UGC. Extensive ablations also prove the effectiveness of our method.
\end{itemize}

\begin{figure*}[t]
     \centering
     \begin{subfigure}[b]{0.24\textwidth}
         \centering
         \includegraphics[width=\textwidth]{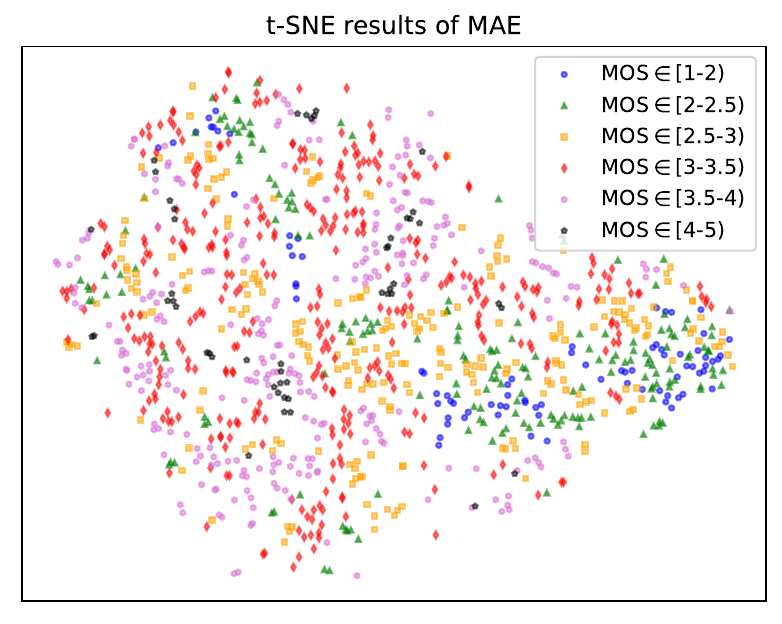}
         \caption{MAE, DBI=8.29}
         \label{fig:mae}
     \end{subfigure}
     \hspace{-0.2cm}
     \begin{subfigure}[b]{0.24\textwidth}
         \centering
         \includegraphics[width=\textwidth]{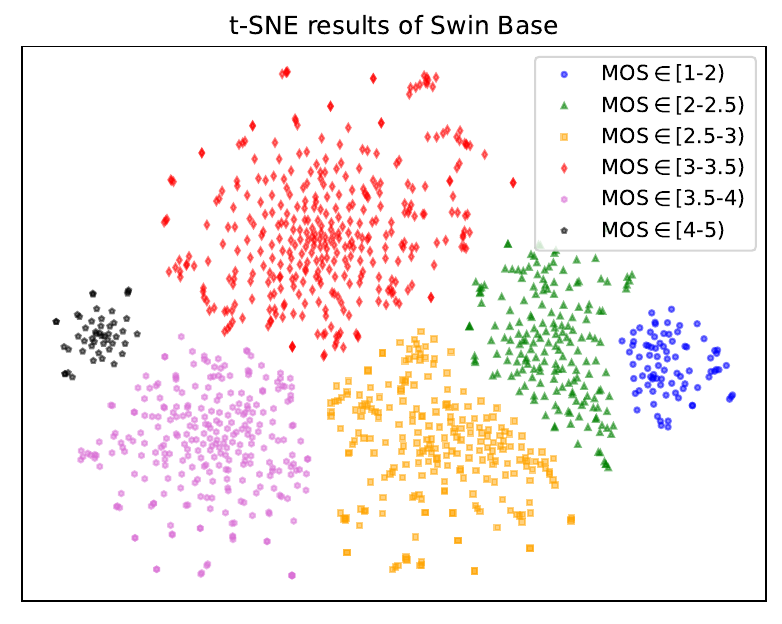}
         \caption{Swin-B, DBI=0.72}
         \label{fig:swin_base}
     \end{subfigure}
     \hspace{-0.2cm}
     \begin{subfigure}[b]{0.24\textwidth}
         \centering
         \includegraphics[width=\textwidth]{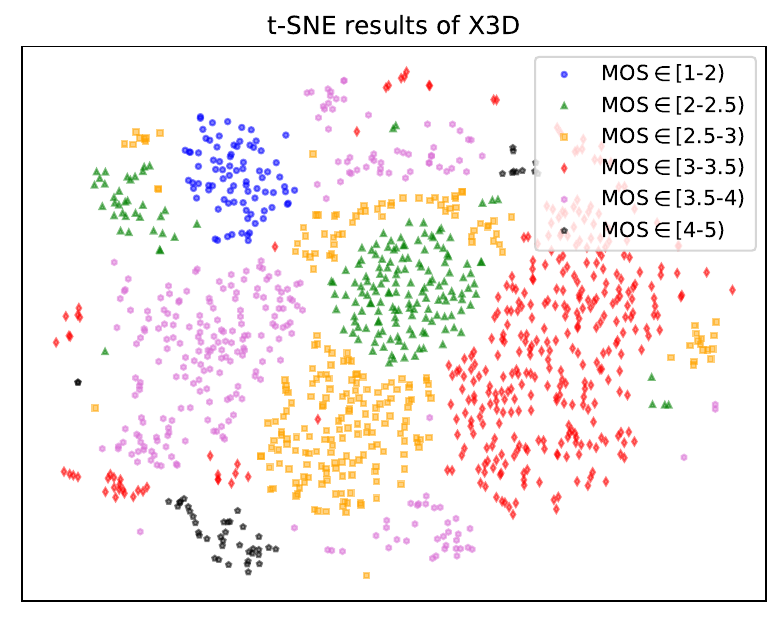}
         \caption{X3D, DBI=2.35}
         \label{fig:x3d}
     \end{subfigure}
     \hspace{-0.2cm}
     \begin{subfigure}[b]{0.24\textwidth}
         \centering
         \includegraphics[width=\textwidth]{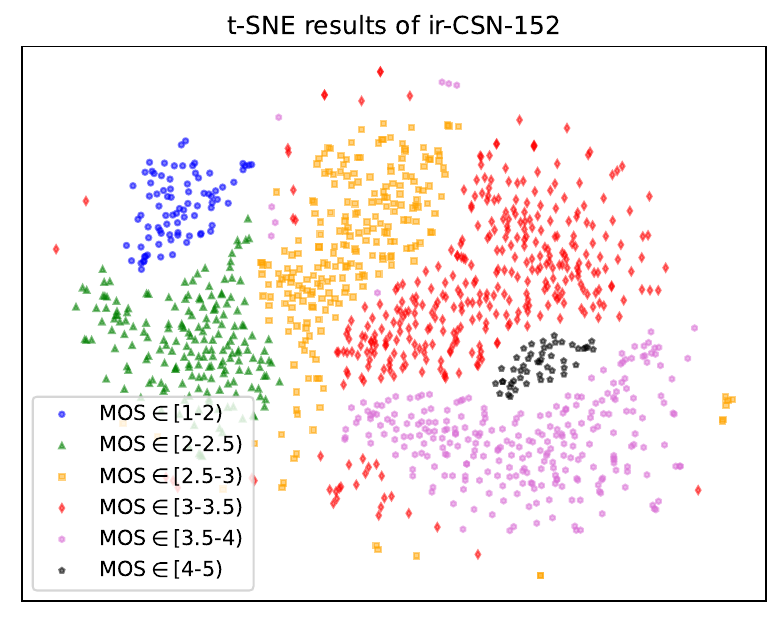}
         \caption{ir-CSN-152, DBI=1.41}
         \label{fig:ir_csn}
     \end{subfigure}
     \hspace{-0.2cm}
     \begin{subfigure}[b]{0.24\textwidth}
         \centering
         \includegraphics[width=\textwidth]{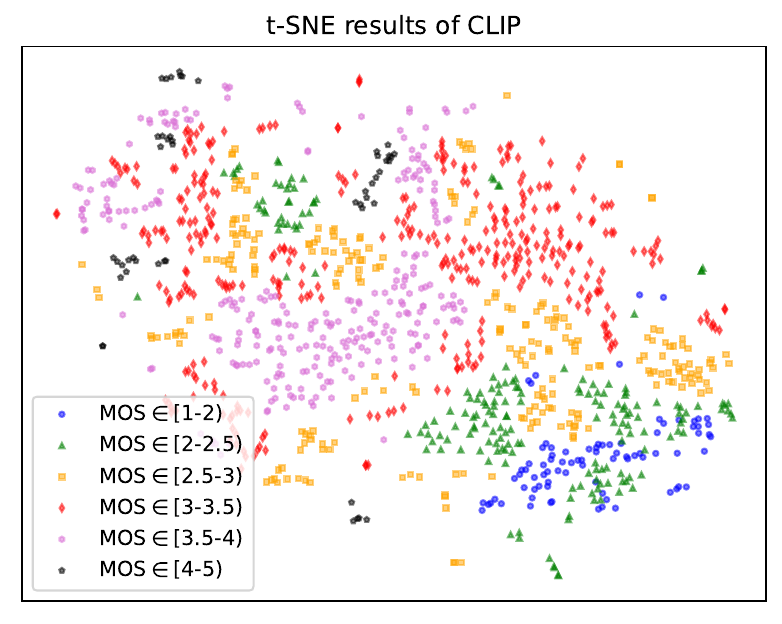}
         \caption{CLIP, DBI=2.49}
         \label{fig:clip}
     \end{subfigure}
     \hspace{-0.2cm}
     \begin{subfigure}[b]{0.24\textwidth}
         \centering
         \includegraphics[width=\textwidth]{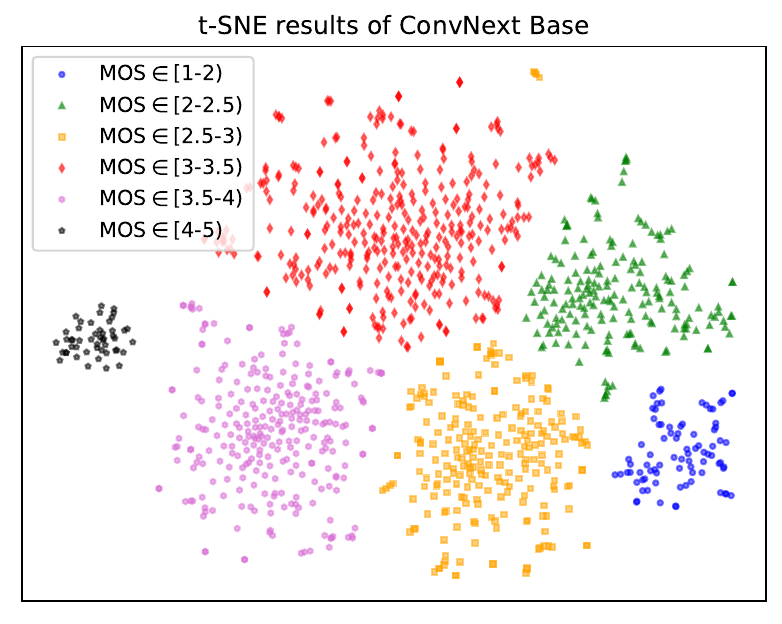}
         \caption{ConvNeXt, DBI=0.62}
         \label{fig:convnext}
     \end{subfigure}
     \hspace{-0.2cm}
     \begin{subfigure}[b]{0.24\textwidth}
         \centering
         \includegraphics[width=\textwidth]{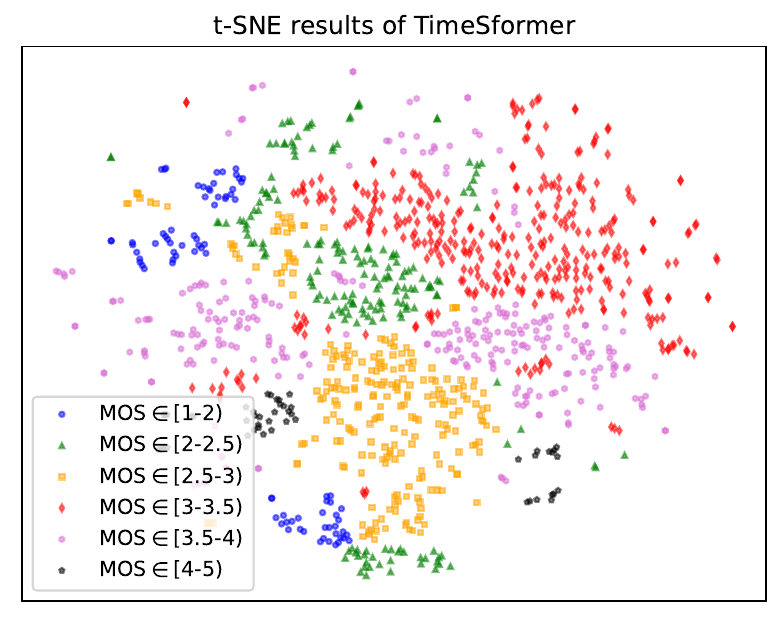}
         \caption{TimeSformer, DBI=4.47}
         \label{fig:timesformer}
     \end{subfigure}
     \hspace{-0.2cm}
     \begin{subfigure}[b]{0.24\textwidth}
         \centering
         \includegraphics[width=\textwidth]{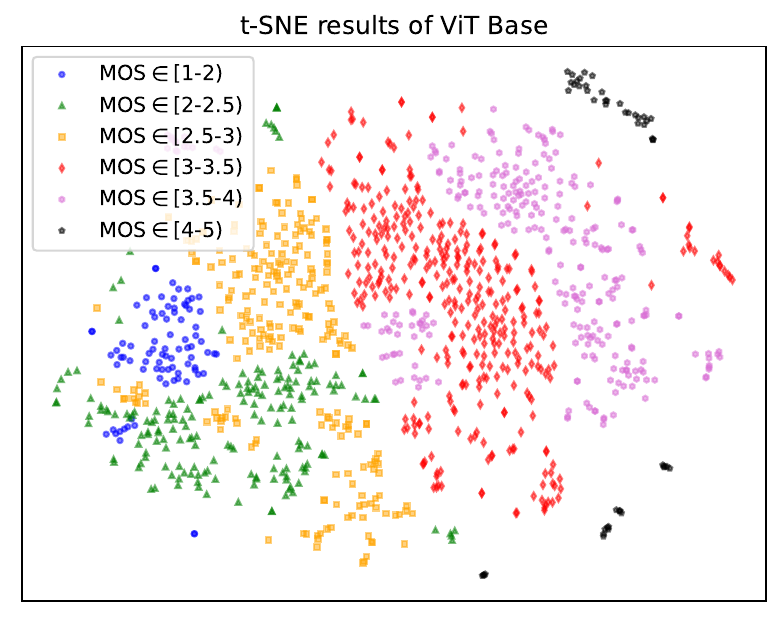}
         \caption{ViT-B, DBI=3.15}
         \label{fig:vit}
     \end{subfigure}
        \caption{Visualization of clustering results of features extracted by different pretrained models using t-SNE \cite{DBLP:journals/jmlr/Maaten09}. Videos in KoNViD-1k \cite{DBLP:conf/qomex/HosuHJLMSLS17} are used. The number of cluster centers is set to be 6 according to the range of MOS values. And DBI scores, which will be introduced in detail in Sec.~ \ref{sec:dbi}, measure the divergence of clustering results (\textbf{the smaller, the better}).
        % The DBI score is also reported for each clustering result. 
        % It can be seen that features generated by some models indicate well discrimination in video quality.
        }
        \label{fig:cluster}
    \vspace{-3mm}
\end{figure*}

\section{Related Work}

\paragraph{VQA.} Based on whether the pristine reference video is required, VQA methods can be classified as Full Reference (FR), Reduced Reference (RR), and No Reference (NR). Our work will be focused on the NR-VQA method.
Traditional NR-VQA methods either measure video quality by rule-based metric \cite{DBLP:journals/spl/YangWCW05}, or predict MOS by an estimator (\eg, Multi-Layer Perceptron, Support Vector Machine) based on hand-crafted features \cite{DBLP:conf/icann/CulibrkKVPZ09}. In recent years, deep learning-based VQA methods have been studied and surpassed traditional methods. 
STDAM \cite{DBLP:conf/mm/XuLZZW021} introduced a graph convolution to extract features and a bidirectional long short-term memory network to handle motion information.
StarVQA \cite{DBLP:journals/corr/abs-2108-09635} proposed encode spatiotemporal
information of each patch on video frames and feed them into a Transformer. RAPIQUE \cite{DBLP:journals/corr/abs-2101-10955} proposed to combine texture features and deep convolutional features. These works, however, neglected the correlation between VQA and other tasks and did not explore other datasets. BVQA \cite{DBLP:journals/tcsv/LiZTZW22} took one step further and proposed to transfer knowledge from IQA and action recognition to VQA. Our work further investigates the possibility of using more kinds of tasks.

\paragraph{Pretrained models.} Pretrained models reveal the great potential in deep learning. In Natural Language Processing (NLP), BERT \cite{DBLP:conf/naacl/DevlinCLT19} and GPT-3 \cite{DBLP:conf/nips/BrownMRSKDNSSAA20}  demonstrated substantial gains on many NLP tasks and benchmarks by pretraining on a large corpus of text followed by finetuning on a specific task. The advent of ViT \cite{DBLP:conf/iclr/DosovitskiyB0WZ21} had migrated this capability into the visual realm. Some subsequent literature \cite{DBLP:conf/icml/RadfordKHRGASAM21,DBLP:journals/corr/abs-2111-06377,DBLP:journals/corr/abs-2111-11429} had shown that the same benefits can be achieved. For example, CLIP \cite{DBLP:conf/icml/RadfordKHRGASAM21} trained on the WebImageText matched the performance of the original ResNet-50 on ImageNet zero-shot, without using any of the original labeled data. In the field of quality assessment (QA), there are also efforts \cite{DBLP:conf/mm/LiJJ19,DBLP:conf/mm/ChenLMWS20,DBLP:journals/tip/MittalMB12,DBLP:journals/tcsv/LiZTZW22} to introduce pretrained models to improve performance. Among them, VSFA \cite{DBLP:conf/mm/LiJJ19} extracted features from a pretrained image classification neural network for its inherent content-aware property. And BVQA \cite{DBLP:journals/tcsv/LiZTZW22} proposed transferring knowledge from IQA and action recognition datasets with motion patterns. Recently, Ada-DQA \cite{DBLP:conf/mm/LiuWYSTZWL23} utilized diverse pretrained models to distill quality-related knowledge. However, its training cost is relatively high. We hope to tap the potential of the pretrained model itself and reduce the tuning process, in this work.

\vspace{-3mm}
\paragraph{Metric learning.} Metric learning can learn distance metrics from data to measure the difference between samples. It has been used in many research, including QA. RankIQA \cite{DBLP:conf/iccv/LiuWB17} trained a siamese network to rank synthesized images with different levels of distortions constrained by pairwise ranking hinge loss and then finetune the model on the target IQA dataset.
UNIQUE \cite{DBLP:journals/tip/ZhangMZY21} sampled ranked image pairs from individual IQA datasets and used a fidelity loss \cite{DBLP:conf/sigir/TsaiLQCM07} and a hinge constraint to supervise the training process.
FPR \cite{DBLP:journals/tip/Chen2022} extracted distortion/reference feature from the input/reference, hallucinated pseudo reference feature from the input alone, and used a triplet loss \cite{DBLP:conf/cvpr/SchroffKP15} to pull the pristine and hallucinated reference features closer while pushing the distortion feature away.
% In our work, we group samples into clusters and propose a centroid triplet loss, trying to pull features of samples within one cluster closer while pushing those from different clusters farther.

\section{Method}

\subsection{Observations}\label{sec:ob}

% Recently, many researches \cite{DBLP:conf/naacl/DevlinCLT19,DBLP:conf/nips/BrownMRSKDNSSAA20,DBLP:conf/icml/RadfordKHRGASAM21,DBLP:journals/corr/abs-2111-06377} are focused on pretraining and show the effectiveness of applying pretrained models to downstream tasks. This meets the main obstacle of VQA tasks, where huge expense of annotating limits the scale of datasets. In the field of VQA, there are also efforts \cite{DBLP:conf/mm/LiJJ19,DBLP:conf/mm/ChenLMWS20,DBLP:journals/tcsv/LiZTZW22} to introduce pretrained models to capture their inherent content-aware properties or motion-related patterns, benefiting the representation of perceptual qualities. However, complicated factors in pretrained models may affect the transferring performance (\eg, architecture of neural networks, pre-text tasks, and pretrained databases). Yet to the best of our knowledge, these factors as well as newly-appeared cutting-edge pretrained models are rarely explored and exploited in the VQA field. So we intend to find a way to make full use of these models.

In recent years, there has been a surge of research attention towards pretraining, as evidenced by a number of notable works \cite{DBLP:conf/naacl/DevlinCLT19,DBLP:conf/nips/BrownMRSKDNSSAA20,DBLP:conf/icml/RadfordKHRGASAM21,DBLP:journals/corr/abs-2111-06377}, that demonstrate the effectiveness of applying pretrained models to downstream tasks. 
This meets the main obstacle of VQA tasks, where the cost of annotation poses a significant challenge in scaling up datasets. In addition to such efforts, the field of VQA has also witnessed endeavors \cite{DBLP:conf/mm/LiJJ19,DBLP:conf/mm/ChenLMWS20,DBLP:journals/tcsv/LiZTZW22} towards leveraging pretraining to capture intrinsic content-aware or motion-related patterns, with a view to enhancing the representation of perceptual qualities. However, the impact of various factors inherent to pretrained models (\eg, neural network architectures, pre-text tasks, and pretrained databases) on the performance of model transfer remains a subject of inquiry. To the best of our knowledge, there has been limited exploration and exploitation of these factors, as well as newly-appeared cutting-edge pretrained models, in VQA. Therefore, our objective is to investigate ways to fully leverage these models in VQA applications.

\begin{figure*}[t]
    \centering
    \includegraphics[width=0.85\linewidth]{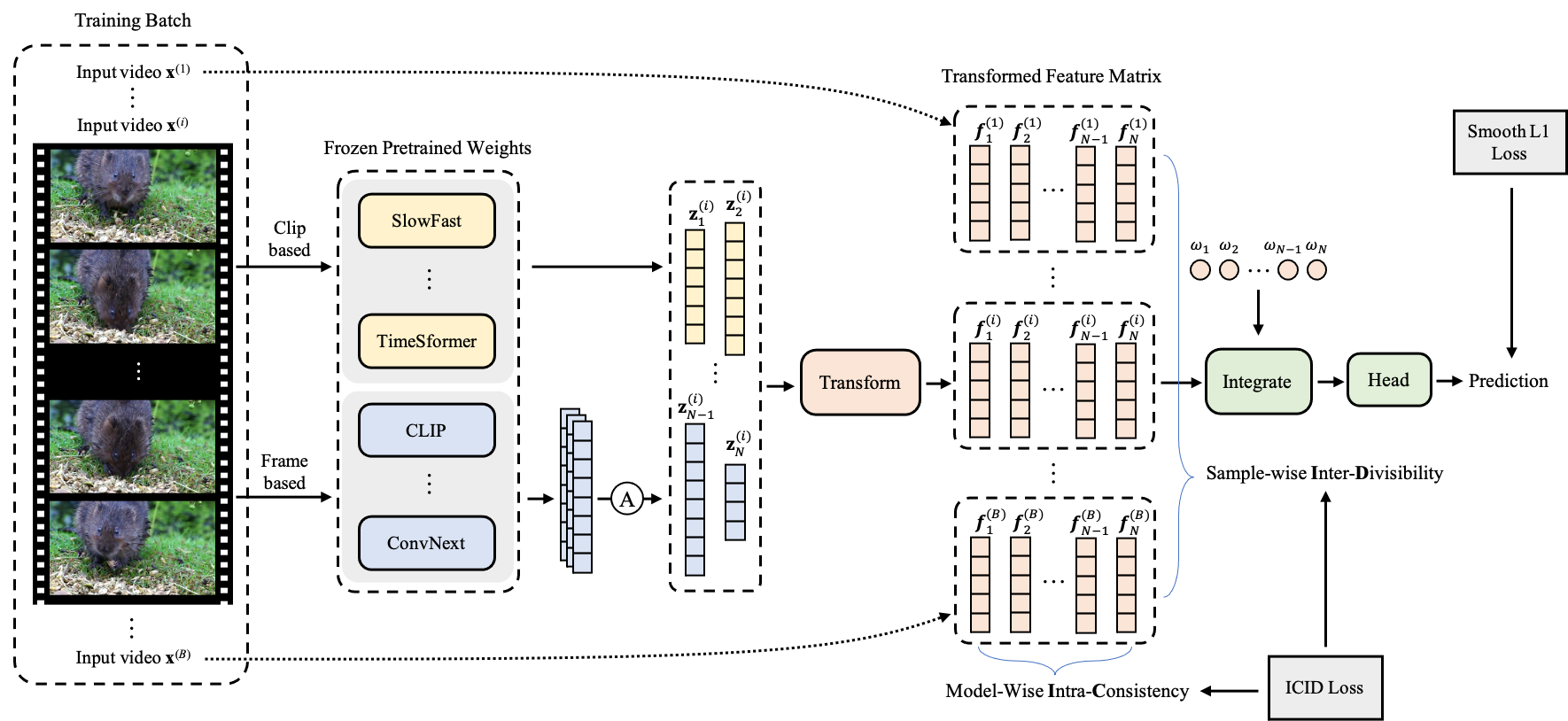}
    \caption{
        The pipeline of the proposed PTM-VQA. Features of input videos are extracted by pretrained models with frozen weights, transformed to the same dimension, and integrated to generate the final representation. Expect for the ordinary smooth $\mathcal{L}_1$ loss for regression, we add an ICID loss to ensure model-wise consistency and sample-wise divisibility.
    }
    \label{fig:framework}
    \vspace{-0.3cm}
\end{figure*}

In order to examine the relationship between pretrained models and VQA tasks, we designed a simple clustering experiment. Specifically, we selected a pretrained model and utilized its frozen weights as a feature extractor to obtain corresponding video features. We then clustered these features into multiple centers using LMNN \cite{DBLP:journals/jmlr/WeinbergerS09}, based on their range of MOS values. To this end, we selected eight models, including MAE \cite{DBLP:journals/corr/abs-2111-06377} trained on ImageNet-1k \cite{DBLP:conf/cvpr/DengDSLL009}, Swin-Base \cite{DBLP:conf/iccv/LiuL00W0LG21} trained on ImageNet-22k \cite{DBLP:conf/cvpr/DengDSLL009}, X3D \cite{DBLP:conf/cvpr/Feichtenhofer20} trained on Kinetics-400 \cite{DBLP:journals/corr/KayCSZHVVGBNSZ17}, ir-CSN-152 \cite{DBLP:conf/iccv/TranWFT19} trained on Sports-1M \cite{KarpathyCVPR14/sports-1m}, CLIP \cite{DBLP:conf/icml/RadfordKHRGASAM21} trained on WebImageText \cite{DBLP:conf/icml/RadfordKHRGASAM21}, ConvNeXt \cite{DBLP:journals/corr/abs-2201-03545} trained on ImageNet-22k, TimeSformer \cite{DBLP:conf/icml/BertasiusWT21} trained on Kinetics-400, and ViT-Base \cite{DBLP:conf/iclr/DosovitskiyB0WZ21} trained on ImageNet-22k. In Fig.~\ref{fig:cluster}, \textbf{we show that some of these models display surprisingly discriminatory results, despite not having been exposed to quality-related labels during pre-text task training}. We hypothesize that some quality-aware representations were learned concurrently during the pre-text task training. For instance, CLIP, which learns visual concepts through natural language supervision, may include emotional descriptions relating to image quality in some texts. Similarly, other models trained on action tasks (\eg, ir-CSN-152) may be sensitive to motion-related distortions \cite{DBLP:journals/tcsv/LiZTZW22} (\eg, camera shaking or motion blurriness). As such, these broader pretrained models may be useful in improving VQA task performance.

\subsection{Pipeline of the proposed PTM-VQA}\label{sec:pipeline}

% Suppose there exist multiple available pretrained models, the most intuitive way to apply them to VQA tasks is finetuning on target datasets and integrating extracted features for quality prediction. Nevertheless, this is highly computationally resource-consuming and becomes less practical as the number of pretrained models increases and the models get larger. For example, the training of ViT \cite{DBLP:conf/iclr/DosovitskiyB0WZ21} requires a TPUv3 with 8 cores in 30 days. And MAE \cite{DBLP:journals/corr/abs-2111-06377} consumes 128 TPUv3 cores for its 800-epoch training. This would be unaffordable in a VQA task. Fortunately, the above results in \ref{fig:cluster} suggest that pretrained models can be applied directly with their weights frozen. In this paper, we propose a simple framework, named PTM-VQA, to utilize the knowledge from diverse pretrained models efficiently. 

Assuming the availability of multiple pre-trained models, the conventional approach for employing them in VQA tasks is through fine-tuning on target datasets while integrating extracted features for quality prediction. Nonetheless, this approach is computationally resource-intensive, making it less feasible as the number of pre-trained models increases and their sizes become larger. For instance, the ViT model requires a TPUv3 with eight cores and 30 days of training \cite{DBLP:conf/iclr/DosovitskiyB0WZ21}, while the MAE model consumes 128 TPUv3 cores and 800 epochs of training \cite{DBLP:journals/corr/abs-2111-06377}. This would be unaffordable in a VQA task. However, the findings illustrated in Fig.~\ref{fig:cluster} suggest that pretrained models have the potential to be applied to VQA tasks with their weights frozen. In this paper, we propose an effective framework, named PTM-VQA, to utilize the knowledge from diverse pretrained models efficiently. 

As shown in Fig.~\ref{fig:framework}, given an input video $\mathbf{x}^{(i)}$, $N$ pretrained models, whose weights are frozen, are utilized to extract features, resulting in representations from different perspectives. Specifically, for video clip-based models, we uniformly sample frames in the temporal dimension to form the input clip. Corresponding representations are then generated by these models. For frame-based models, they are fed with sampled frames and the output features are averaged to perform the spatiotemporal representation. Features extracted by models can be noted as $\mathbf{z}_n^{(i)}$, where $n\in\{1,\dots,N\}$. To further distill quality-aware features and perform dimension alignment, we apply a learnable transformation module following each feature extractor. Structurally, the transformation module consists of two fully connected layers, each followed by a normalization layer and an activation layer of GELU. The transformed features are defined as $\mathbf{f}_n^{(i)}\in \mathbb{R}^{D}$, where $D$ represents the aligned dimension. Then features are integrated to obtain a unified representation through:
\begin{equation}\label{eq:agg}
    \mathbf{h}^{(i)}=\frac{\sum^{N}_{n=1}\omega_n\mathbf{f}^{(i)}_{n}}{\sum^{N}_{n=1}\omega_n},
\end{equation}
where $\omega_n$ is the coefficient for each model. When $\omega_n$ equals $1/N$, it means calculating an average, with each model contributing equally to the final representation. Last, $\mathbf{h}^{(i)}$ is used to get the quality prediction through a regression head, which is a single fully-connected layer.

\begin{figure*}[t]
    \centering
    \includegraphics[width=0.85\linewidth]{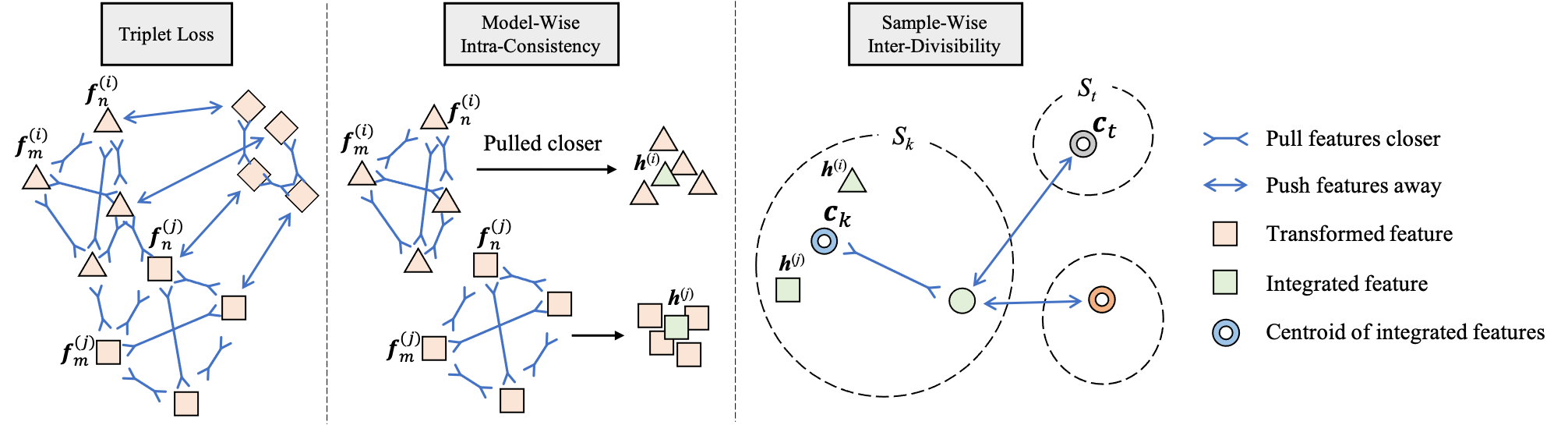}
    \caption{
    Illustration of ICID loss. The figure shows examples of several triplets of triplet loss; two sets of intra-consistency between features extracted by four pretrained models; and one sample (with two triplets) for inter-divisibility.
    % Illustration of the proposed consistency loss and centroid triplet loss. The figure shows examples of two samples for consistency loss, and one sample (with two triplets) for centroid triplet loss. Please note that only some of the features are drawn for the sake of simplification.
    }
    \label{fig:loss}
    \vspace{-0.3cm}
\end{figure*}

Drawing on the proposed design, the training methodology exhibits remarkable efficiency, thereby circumventing the computational overhead associated with the finetuning approach aforementioned. As attested by the results presented in Tab.~\ref{tab:training_detail}, the entire training regimen can be accomplished within a span of approximately \textbf{two hours, leveraging a single GPU}. This retains the information of the pretrained models well, but it also increases the difficulty of obtaining preferable performance due to the reduction of learnable parameters. Some concerns are as follows:
\begin{itemize}
    \item[1.] Due to various pre-texts of pretrained models, features generated by different models are of \textbf{large diversity}, which may distribute over inconsistent feature spaces \cite{DBLP:conf/icml/WortsmanIGRLMNF22}. How to constrain these abundant features into a unified quality-aware space is important.
    \item[2.] Different from the objective category in common classification tasks, the perceptual quality of a video is more implicit and related to various factors (\eg, content attractiveness, distortion type, and level, motion pattern and level), whereas videos of the same quality often render completely different content and vice versa. Therefore, it is difficult for the models trained based on objective annotations to distinguish these samples of \textbf{the same category} but with a large perceptual \textbf{quality difference}. A more comprehensive contrast approach beyond sample-wise comparison needs to be proposed to deal with these outliers.
    \item[3.] There exist \textbf{hundreds of} pretrained models available in public libraries. How to select the desired models efficiently and how to determine the contribution of these models to represent the perceptual quality effectively is an urgent problem to be solved.
\end{itemize}

\subsection{Intra-Consistency and Inter-Divisibility Loss}

To solve the above concerns and better satisfy VQA tasks, we intend to constrain the features between different pretrained models and different samples using metric learning. Triplet loss, which is one of the most widely adopted metric learning measures, can be formed as follows:
\begin{equation}\label{equ:triplet}
\small
    \mathcal{L}_{\text{triplet}}
    (\mathbf{f}_{\hat{a}},\mathbf{f}_{\hat{p}},\mathbf{f}_{\hat{n}}) = \max(\|\mathbf{f}_{\hat{a}}-\mathbf{f}_{\hat{p}}\|^2-
    \|\mathbf{f}_{\hat{a}}-\mathbf{f}_{\hat{n}}\|^2+\alpha,0),
\end{equation}
where $\mathbf{f}_{\hat{a}}$,$\mathbf{f}_{\hat{p}}$,$\mathbf{f}_{\hat{n}}$ are features of an anchor sample $\hat{a}$, a positive sample $\hat{p}$ of the same class as $\hat{a}$, and a negative sample $\hat{n}$ which has a different class of $\hat{a}$. And $\alpha$ is a margin between anchor-positive and anchor-negative pairs. Some previous studies \cite{DBLP:journals/tip/Chen2022,Golestaneh_2022_WACV} in QA also applied triplet loss to measure the distance between the distorted feature and the reference feature of the same sample. Since the MOS values are continuous, the original triplet loss cannot be directly used to constrain the distance between arbitrary samples. We make some modifications to constrain features generated by different pretrained models and samples, as given in Fig.~\ref{fig:loss}.

\paragraph{Intra-consistency constraint.} \textbf{To solve the first concern}, and unify features generated by different pretrained models into a unified quality-aware latent space, we propose a model-wise intra-consistency constraint. Formally, it is defined to minimize the distance between arbitrary two of the transformed features through computing a cosine similarity, which is widely used in deep metric learning \cite{DBLP:conf/cvpr/WangHHDS19}:
\begin{equation}
\small
    \mathcal{L}_{\text{intra}} = \frac{2}{N\cdot(N-1)} \sum_{n=1}^{N} \sum_{m, m\neq n}^N \big(1- 
    \frac{\mathbf{f}^{(i)}_n \cdot \mathbf{f}^{(i)}_m}{\|\mathbf{f}^{(i)}_n\|_2\|\mathbf{f}^{(i)}_m\|_2}
    \big).
\end{equation}

\begin{table*}[t]
% \caption{Training details of PTM-VQA. By using pretrained models with frozen weights, the overall computational cost is small.}
\caption{Details of PTM-VQA for different datasets. Time$^1$ is the training time. Time$^2$ represents the inference time using a 1080P/30FPS/20s video. The overall computational cost is relatively small compared with existing SOTA methods.}
% \vspace{-0.2cm}
\scriptsize
\label{tab:training_detail}
    \centering
    \begin{tabular}{c|c|ccc|ccc|cc}
    \toprule
    Dataset & Combination of Pretrained Models & Frames & Interval & LR & Param(M) & FLOPs(T) & Mem(G) & Time$^1$(h) & Time$^2$(s) \\
    \midrule
    KoNViD-1k   & ConvNeXt, ir-CSN-152, CLIP               & 16 & 2 & 1e-3 & 0.66 & 0.43 & 4.94 & 2.00 & 0.14 \\
    LIVE-VQC    & CLIP, TimeSformer                        & 16 & 4 & 5e-3 & 0.30 & 0.53 & 4.32 & 1.97 & 0.17 \\
    YouTube-UGC & ConvNeXt, ir-CSN-152, CLIP, Video Swin-B & 32 & 8 & 1e-3 & 0.86 & 1.35 & 5.32 & 2.34 & 1.20 \\
    \bottomrule
    \end{tabular}
    \vspace{-0.2cm}
\end{table*}

\paragraph{Inter-divisibility constraint.} \textbf{To solve the second concern}, we split videos into distinct pseudo clusters under different numerical intervals, according to the annotated MOS values (on a scale of 1.0 to 5.0). For example, videos with MOS in the range of 1.0 to 2.0 are generally considered to be of poor quality, and whose content cannot be normally recognized due to the existence of various distortions. And videos with MOS in the range of 4.0 to 5.0 are of high quality, whose content is unambiguous, without noise, shaking, and blurring. We identify the videos within the same range as the same category, thus dividing them into $K$ clusters. Each cluster can be noted as $\mathcal{S}_k = \{\mathbf{x}^{(i)}|y^{(i)}\in(p_k, q_k],q_k>p_k\in[1.0,5.0]\}$, where $y^{(i)}$ is the labeled MOS for the $i$-th input video, $p_k$ and $q_k$ are the endpoints of the interval. Through this pseudo cluster, triplet loss can be utilized for samples belonging to the same cluster to be closer and samples of the different clusters to be farther away. Then Equ.~\ref{equ:triplet} can be rewritten as:
\begin{equation}
\small
    \mathcal{L}_{\text{triplet}}(\mathbf{h}^{(i)},\mathbf{h}^{(j)},\mathbf{h}^{(l)}), ~ \text{where} ~ \mathbf{x}^{(i)},\mathbf{x}^{(j)} \in \mathcal{S}_k, \mathbf{x}^{(l)} \notin \mathcal{S}_k.
\end{equation}
Besides, the original feature $\mathbf{f}$ extracted by individual models is replaced by the integrated feature $\mathbf{h}$.
As shown in Fig.~\ref{fig:loss}, the original triplet loss performs a sample-to-sample form, which is highly affected by the sampling of triples. When facing outliers that are of the same quality but render different contents or vice versa, it may lead to bad local minima and prevent the model from achieving top performance. Thus we propose using the centroid of the cluster to represent the positive and negative points as:
\begin{equation}
\small
\begin{aligned}
    & \mathcal{L}_{\text{inter}} = \mathcal{L}_{\text{triplet}}(\mathbf{h}^{(i)},\mathbf{c}_{k},\mathbf{c}_{t}), ~ \text{where} \\
    & \mathbf{c}_{k} = \frac{1}{|\mathcal{S}_k|}\sum_{\{i|\mathbf{x}^{(i)}\in\mathcal{S}_k\}} \mathbf{h}^{(i)}, 
    \mathbf{c}_{t} = \frac{1}{|\mathcal{S}_t|}\sum_{\{j|\mathbf{x}^{(j)}\in\mathcal{S}_t\}} \mathbf{h}^{(j)}.
\end{aligned}
\end{equation}

Given a batch consisting $B$ inputs, during training, the optimization objective can be summarized as:
\begin{equation}
\small
    \text{min}~ \mathcal{L}_1 + \beta \big(\sum\nolimits^{B}_{i=1}\mathcal{L}_{\text{intra}} + \mathcal{L}_{\text{inter}}\big),
\end{equation}
where $\beta$ is the coefficient balancing smooth $\mathcal{L}_1$ regression loss and the proposed ICID loss.

\subsection{Selection scheme through DBI}\label{sec:dbi}

\textbf{For the third concern}, we observe an obvious difference in the clustering results of different pretrained models in Fig.~\ref{fig:cluster}. Since the weights of models are frozen both in the clustering test and subsequent training process, the divergence of clustering results can reflect the relevance of VQA tasks. We propose using the Davies-Bouldin Index (DBI) \cite{DBLP:journals/pami/DaviesB79} as a metric for model selection, which is commonly employed for evaluating clustering results. In our particular setting, the DBI can be expressed as follows:
\begin{equation}
\small
\begin{aligned}
    & \psi = \frac{1}{K}\sum^{K}_{k=1}\max\limits_{t\neq k}\frac{
    \mathbf{d}_k + \mathbf{d}_t}{
    \|\mathbf{c}_{k}-\mathbf{c}_{t}\|_2}, ~\text{where} \\
    & \mathbf{c}_{k} = \frac{1}{|\mathcal{S}_k|} \sum_{\mathcal{S}_k}\mathbf{z}^{(i)},
    \mathbf{d}_k = \frac{1}{|\mathcal{S}_k|}
    \sum_{\mathcal{S}_k}\|\mathbf{z}^{(i)}-\mathbf{c}_k\|_2,
\end{aligned}
\end{equation}
where $\mathbf{c}_k$ is the centroid of cluster $\mathcal{S}_k$ for the set of extracted feature $\mathbf{z}^{(i)}$, $d_k$ represents the average distance between each sample and its corresponding centroid. For the $n$-th model, its DBI score can be noted as $\psi_n$. A lower DBI indicates better clustering performance, which means that the pretrained model (\eg, ConvNeXt, Swin-Base, ir-CSN-152, CLIP in Fig.~\ref{fig:cluster}) is more relevant to downstream VQA tasks. During training, the DBI scores computed offline can be used in the aggregation procedure as given in Equ.~\ref{eq:agg}, where $\omega_n$ can be replaced by $1/\psi_n$. It means the models that are more relevant to the VQA task contribute more to the feature representation.

\section{Experiments}

\subsection{Datasets and evaluation criteria}

\textbf{Datasets.} Our method is evaluated on 4 public NR-VQA datasets, including KoNViD-1k \cite{DBLP:conf/qomex/HosuHJLMSLS17}, LIVE-VQC \cite{DBLP:journals/tip/SinnoB19/live-vqc}, YouTube-UGC \cite{DBLP:conf/mmsp/WangIA19/youtube-ugc} and LSVQ \cite{DBLP:conf/cvpr/YingMGB21}. In detail, KoNViD-1k contains 1,200 videos that are fairly filtered from a large public video dataset YFCC100M. The videos are 8 seconds long with 24/25/30 FPS and a resolution of $960 \times 540$. The MOS ranges from 1.22 to 4.64. Each video owns 114 annotations to get a reliable MOS. LIVE-VQC consists of 585 videos with complex authentic distortions captured by 80 different users using 101 different devices, with 240 annotations for each video. YouTube-UGC has 1,380 UGC videos sampled from YouTube with a duration of 20 seconds and resolutions from 360P to 4K, with 123 annotations for each video. 
And LSVQ is the largest VQA dataset currently (proposed in 2021) with 39,076 videos.
All the datasets contain no pristine videos, thus only NR methods can be evaluated on them. Following \cite{DBLP:conf/mm/XuLZZW021,DBLP:conf/cvpr/SuYZZGSZ20}, we split the dataset into a 80\% training set and a 20\% testing set randomly for the first three datasets. For LSVQ, we follow the official split setting.  We perform 10 repeat runs in each dataset using different splittings to get the mean values of PLCC and SRCC.

\textbf{Evaluation criteria.} Pearson’s Linear Correlation Coefficient (PLCC) and Spearman’s Rank-Order Correlation Coefficient (SRCC) are selected as criteria to measure the accuracy and monotonicity. They are in the range of $\displaystyle [0, 1]$. A larger PLCC means a more accurate numerical fit with MOS scores. A larger SRCC shows a more accurate ranking between samples. Besides, the mean average of PLCC and SRCC is also reported as a comprehensive criterion.

\begin{table*}[t]
\caption{
    Comparisons with existing methods. The ``-" is an unreported result. The ``*" is using extra labeled data for training. The best and second best results are \textbf{bolded} and \underline{underlined}. Average scores weighted by the size of datasets are also reported.
    % PTM-VQA obtains SOTA results in three datasets with rather a small amount of learnable weights.
    }
    \label{tab:sota_performance}
    \centering
    \footnotesize
    \begin{tabular}{c|ccc|ccc|ccc|ccc}
    \toprule
    \multirow{2}{*}{Method} & \multicolumn{3}{c|}{KoNViD-1k} & \multicolumn{3}{c|}{LIVE-VQC} & \multicolumn{3}{c|}{YouTube-UGC} & \multicolumn{3}{c}{Weighted Avg}  \\
    & PLCC & SRCC & Mean & PLCC & SRCC & Mean & PLCC & SRCC & Mean & PLCC & SRCC & Mean \\
    \midrule
    VIIDEO \cite{DBLP:journals/tip/MittalSB16} & 0.3030 & 0.2980 & 0.3005 & 0.2164 & 0.0332 & 0.1248 & 0.1534 & 0.0580 & 0.1057 & 0.2218 & 0.1444 & 0.1832 \\
    NIQE \cite{DBLP:journals/spl/MittalSB13/niqe} & 0.5530 & 0.5417 & 0.5473 & 0.6286 & 0.5957 & 0.6121 & 0.2776 & 0.2379 & 0.2577 & 0.4469 & 0.4365 & 0.4417 \\
    BRISQUE \cite{DBLP:journals/tip/MittalMB12} & 0.626 & 0.654 & 0.640 & 0.638 & 0.592 & 0.615 & 0.395 & 0.382 & 0.388 & 0.5275 & 0.5240 & 0.5257 \\ 
    VSFA \cite{DBLP:conf/mm/LiJJ19} & 0.744 & 0.755 & 0.749 & - & - & - & - & - & - & - & - & - \\
    TLVQM \cite{DBLP:journals/tip/Korhonen19} & 0.7688 & 0.7729 & 0.7708 & 0.8025 & 0.7988 & 0.8006 & 0.6590 & 0.6693 & 0.6641 & 0.7272 & 0.7325 & 0.7298 \\
    RIRNet \cite{DBLP:conf/mm/ChenLMWS20} & 0.7812 & 0.7755 & 0.7783& 0.7982 & 0.7713 & 0.7847 & - & - & - & - & - & - \\
    UGC-VQA \cite{DBLP:journals/tip/TuWBAB21} & 0.7803 & 0.7832 & 0.7817 & 0.7514 & 0.7522 & 0.7518 & 0.7733 & 0.7787 & 0.7760 & 0.7663 & 0.7732 & 0.7697 \\
    CSPT \cite{DBLP:journals/tip/ChenLWDS22} & 0.8062 & 0.8008 & 0.8035 & 0.8194 & 0.7989 & 0.8091 & - & - & - & - & - & - \\
    RAPIQUE \cite{DBLP:journals/corr/abs-2101-10955} & 0.8175 & 0.8031 & 0.8103 & 0.7863 & 0.7548 & 0.7705 & 0.7684 & 0.7591 & 0.7637 & 0.7925 & 0.7789 & 0.7857 \\
    StarVQA \cite{DBLP:journals/corr/abs-2108-09635} & 0.796 & 0.812 & 0.804 & 0.808 & 0.732 & 0.770 & - & - & - & - & - & - \\
    BVQA* \cite{DBLP:journals/tcsv/LiZTZW22} & 0.8335 & 0.8362 & 0.8348 & {0.8415} & \textbf{0.8412} & \textbf{0.8413} & 0.8194 & 0.8312 & 0.8253 & {0.8352} & {0.8349} & {0.8351} \\
    STDAM* \cite{DBLP:conf/mm/XuLZZW021} & {0.8415} & {0.8448} &{0.8431} & {0.8204} & 0.7931 & 0.8067 & {0.8297} & {0.8341} & {0.8319} & 0.8325 & 0.8337 & 0.8331 \\
    Fast-VQA \cite{DBLP:conf/eccv/WuCHLWSYL22} & {0.855} & \textbf{0.859} & {0.857} & \textbf{0.844} & \underline{0.823} & \underline{0.834} & - & - & - & - & - & - \\
    VQT \cite{DBLP:conf/mm/YuanKZSW23} & \underline{0.8684} & \underline{0.8582} & \underline{0.8633} & \underline{0.8357} & 0.8238 & 0.8298 & \underline{0.8514} & \underline{0.8357} & \underline{0.8436} & \underline{0.8529} & \underline{0.8421} & \underline{0.8475} \\
    \midrule
    PTM-VQA & \textbf{0.8718} & {0.8568} & \textbf{0.8643} & 0.8198 & {0.8110} & {0.8154} & \textbf{0.8570} & \textbf{0.8578} & \textbf{0.8574} & \textbf{0.8591} & \textbf{0.8454} & \textbf{0.8523} \\
    \bottomrule
    \end{tabular}
\end{table*}

\subsection{Implementation details}\label{sec:exp_detail}

Our experiments are performed using PyTorch \cite{DBLP:conf/nips/PaszkeGMLBCKLGA19} and MMAction2 \cite{2020mmaction2}, and are all conducted on \textbf{one} Nvidia V100 GPU by training for 60 epochs. For KoNViD-1k, we select ConvNeXt, ir-CSN-152, and CLIP as feature extractors. For LIVE-VQC, we use CLIP and TimeSformer. For YouTube-UGC, an extra Video Swin-Base is used together with those selected on KoNViD-1k. For KoNViD-1k, we sample 16 frames with a frame interval of 2. As videos in LIVE-VQC and YouTube-UGC has a longer time duration, we use larger intervals for these two datasets. Since most augmentations will introduce extra interference to the quality of videos \cite{DBLP:conf/iccv/KeWWMY21}, we only choose the center crop to produce an input with a size of $224 \times 224$. During training, we use AdamW optimizer with a weight decay of 0.02. Cosine annealing with a warmup of 2 epochs is adopted to control the learning rate. The dimension $D$ of transformed features is set to 128. The margin $\alpha$ is set to be 0.05. $\beta$ is set to be 0.2. By default, we select the checkpoint generated by the last iteration for evaluation. During inference, we follow a similar procedure as given in \cite{DBLP:journals/corr/abs-2103-15691/vivit} by using $4\times5$ views. To be specific, 4 clips are uniformly sampled from a video in the temporal domain. For each clip, we take 5 crops in the four corners and the center. The final score is computed as the average score. More details are given in Tab.~\ref{tab:training_detail}.

\subsection{Comparison with SOTA methods}

\begin{table}[t]
    \centering
    \footnotesize
    \caption{Comparisons on the largest LSVQ dataset.}
    \begin{tabular}{c|cc|cc}
    \toprule
        \multirow{2}{*}{Method} & \multicolumn{2}{c|}{LSVQ-Test} & \multicolumn{2}{c}{LSVQ-1080P} \\
        & PLCC & SRCC & PLCC & SRCC \\
    \midrule
    BRISQUE \cite{DBLP:journals/tip/MittalMB12}	    & 0.576 & 0.569 & 0.531 & 0.497 \\
    VSFA \cite{DBLP:conf/mm/LiJJ19}                 & 0.796 & 0.801 & 0.704 & 0.675 \\
    TLVQM  \cite{DBLP:journals/tip/Korhonen19}      & 0.774 & 0.772 & 0.616 & 0.589 \\
    PVQ (w/o patch) \cite{DBLP:conf/cvpr/YingMGB21} & 0.816 & 0.814 & 0.708 & 0.686 \\
    PVQ (w patch)   \cite{DBLP:conf/cvpr/YingMGB21} & 0.828 & 0.827 & 0.739 & 0.711 \\
    \midrule
    PTM-VQA-1k      & 0.8536 & 0.8530 & 0.7784 & 0.7279 \\
    PTM-VQA-VQC     & \textbf{0.8637} & \textbf{0.8545} & \textbf{0.7817} & \textbf{0.7359} \\
    PTM-VQA-UGC     & 0.8443 & 0.8429 & 0.7769 & 0.7300 \\
    \bottomrule
    \end{tabular}
    \label{tab:lsvq}
\end{table}

We select existing VQA methods for comparison in three datasets. As shown in Tab.~\ref{tab:sota_performance}, our method obtains competitive results on all three datasets. Compared with traditional methods that rely on statistical regularities (\eg, VIIDEO \cite{DBLP:journals/tip/MittalSB16}, NIQE \cite{DBLP:journals/spl/MittalSB13/niqe}, and BRISQUE \cite{DBLP:journals/tip/MittalMB12}), PTM-VQA models outperform by large margins. Compared with some deep learning-based methods that apply well-designed networks (\eg, TLVQM \cite{DBLP:journals/tip/Korhonen19}, StarVQA \cite{DBLP:journals/corr/abs-2108-09635}), PTM-VQA still obtains higher performances. Especially, VSFA \cite{DBLP:conf/mm/LiJJ19} and RIRNet \cite{DBLP:conf/mm/ChenLMWS20} also adopt pretrained models that contain content-dependency or motion information to finetune in VQA tasks. PTM-VQA demonstrates that features extracted directly from pretrained models can also achieve better results. As the best two SOTA methods BVQA \cite{DBLP:journals/tcsv/LiZTZW22} and STDAM \cite{DBLP:conf/mm/XuLZZW021} who utilize extra IQA datasets, \textbf{PTM-VQA proves that transferring knowledge from pretrained models can achieve competitive results compared with a model trained with additional data}. 
% PTM-VQA improved SOTA's PLCC by 1.7\% and mean score by 0.7\% on KoNViD-1k. And PTM-VQA improved SOTA's PLCC by 2.73\%, SRCC by 2.37\%, and mean score by 2.55\% on YouTube-UGC. 

To assess the generalizability of the selected combinations, we evaluate on the largest LSVQ dataset using the three combinations utilized in KoNViD-1k, LIVE-VQC, and YouTube-UGC. As given in Tab.~\ref{tab:lsvq}, PTM-VQA models demonstrate a significant performance advantage over existing methods, \textbf{indicating the benefits of leveraging pretrained models with a larger amount of data}.

We also compare the cost of inference time with open-source methods on a 1080P video (100 repeat runs). And the inference time cost is 75s (BRISQUE), 248s (TLVQM), 117s (VSFA), 0.12s (StarVQA), and 2.45s (BVQA) respectively. Thanks to the \textbf{reduced dimensions (\eg, frame sampling, center cropping) and model selection using DBI}, PTM-VQA models do not significantly increase inference time over StarVQA and BVQA, as given in Tab.~\ref{tab:training_detail}. Meanwhile, due to the different number and composition of pretrained models, the calculation cost of PTM-VQA models vary. Even so, the largest PTM-VQA can process a high-resolution video in about 1s, and the smaller models can process nearly 6/7 videos per second. 
% In future work, we consider speeding up the inference using methods such as knowledge distillation.

\paragraph{Cross-database comparison.} To emphasize the validity and generalizability, we perform the cross-database evaluation in Tab.~\ref{tab:corss-dataset}. Models trained on LSVQ are tested on much smaller datasets of KoNViD-1k and LIVE-VQA directly. It can be seen that PTM-VQA transferred very well to both datasets, highlighting the general efficacy.

\begin{table}[t]
    \centering
    \footnotesize
    \caption{Cross-database comparison using models trained on LSVQ, then evaluated on the entire KoNViD-1k and LIVQ-VQC datasets without fine-tuning.}
    \vspace{-3mm}
    \begin{tabular}{c|cc|cc}
    \toprule
        \multirow{2}{*}{Method} & \multicolumn{2}{c|}{KoNViD-1k} & \multicolumn{2}{c}{LIVQ-VQC} \\
        & PLCC & SRCC & PLCC & SRCC \\
    \midrule
    BRISQUE \cite{DBLP:journals/tip/MittalMB12}	    & 0.647 & 0.646 & 0.536 & 0.524 \\
    VSFA \cite{DBLP:conf/mm/LiJJ19}                 & 0.794 & 0.784 & 0.772 & 0.734 \\
    TLVQM  \cite{DBLP:journals/tip/Korhonen19}      & 0.724 & 0.732 & 0.691 & 0.670 \\
    PVQ (w/o patch) \cite{DBLP:conf/cvpr/YingMGB21} & 0.781 & 0.781 & 0.776 & 0.747 \\
    PVQ (w patch)   \cite{DBLP:conf/cvpr/YingMGB21} & 0.795 & 0.791 & \textbf{0.807} & \textbf{0.770} \\
    \midrule
    PTM-VQA-1k      & 0.8249 & 0.8164 & 0.7772 & 0.7215 \\
    PTM-VQA-VQC     & 0.8292 & \textbf{0.8243} & 0.7772 & 0.7231 \\
    PTM-VQA-UGC     & \textbf{0.8303} & 0.8223 & 0.7852 & 0.7367 \\
    \bottomrule
    \end{tabular}
    \label{tab:corss-dataset}
\end{table}

\subsection{Ablation studies}\label{sec:ablation}

We conduct experimental analysis to evaluate the effectiveness of each component. By default, experiments are performed following the best configurations in Sec.~\ref{sec:exp_detail}.

\begin{table}[t]
    \centering
    \footnotesize
    \caption{Ablation on the ICID loss in KoNViD-1k.}
    \vspace{-0.3cm}
    \begin{tabular}{cccc|cc}
    \toprule
    $\mathcal{L}_1$ & $\mathcal{L}_{\text{intra}}$ & $\mathcal{L}_{\text{inter}}$ & $\mathcal{L}_{\text{tri}}$ & PLCC & SRCC \\
    \midrule
    \checkmark & \checkmark & \checkmark & & \textbf{0.8718} & \textbf{0.8568} \\
    \midrule
    \checkmark &  &  & \checkmark & 0.7968 & 0.7850 \\
    \checkmark & & & & 0.7867 & 0.7655 \\
    \checkmark & \checkmark & & & 0.8545 & 0.8299 \\
    \checkmark & & \checkmark & & 0.8172 & 0.7707\\
    \bottomrule
    \end{tabular}
    \label{tab:ablation_loss}
    % \vspace{-0.3cm}
\end{table}

\begin{table}[t]
    \centering
    \footnotesize
    \caption{Ablation on cluster settings in KoNViD-1k.}
    \vspace{-3mm}
    \begin{tabular}{c|c|cc}
    \toprule
    $K$ & intervals & PLCC & SRCC \\
    \midrule
    2 & $\mathcal{S}_1$=[1,3), $\mathcal{S}_2$=[3, 5] & 0.8277 & 0.8066 \\
    \midrule
    \multirow{2}{*}{4} & $\mathcal{S}_1$=[1,2), $\mathcal{S}_2$=[2,3), & \multirow{2}{*}{0.8431} & \multirow{2}{*}{0.8012}\\
      & $\mathcal{S}_3$=[3,4), $\mathcal{S}_4$=[4,5]  \\
    \midrule
    \multirow{2}{*}{6} & $\mathcal{S}_1$=[1,2), $\mathcal{S}_2$=[2, 2.5), $\mathcal{S}_3$=[2.5, 3), & \multirow{2}{*}{\textbf{0.8718}} & \multirow{2}{*}{\textbf{0.8568}}  \\
    & $\mathcal{S}_4$=[3, 3.5), $\mathcal{S}_5$=[3.5, 4), $\mathcal{S}_6$=[4, 5] & \\
    \bottomrule
    \end{tabular}
    \label{tab:interval}
    \vspace{-3mm}
\end{table}

\paragraph{Ablation on different constraints.} As given in Tab.~\ref{tab:ablation_loss}, direct usage of triplet loss cannot obtain satisfying results. When either or both constraints are absent, performance degrades significantly. These prove the effectiveness of intra-consistency constraints in transferring knowledge from different pretrained models and inter-divisibility constraints in generating stable predictions.

\paragraph{Ablation on the clustering settings.} Tab.~\ref{tab:interval} gives the results with different numbers of clusters. When $K$ is 2, videos are simply classified as low-quality and high-quality ones. When $K$ is 4, videos are evenly divided into four parts on a scale of 1.0 to 5.0. Due to the relatively small amount of data at both endpoints, a 6-split setting can be obtained by using fine-grained division in the middle fraction segment. Since the need to ensure the number of samples per cluster within the batch, a larger number of clusters are not attempted. The best result can be acquired when $K$ is 6.

% \begin{table}[t]
%     \centering
%     \footnotesize
%     \caption{Correlation of DBIs and performances.}
%     \begin{tabular}{c|c|ccc}
%     \toprule
%     Model & DBI & KoNViD1k & LIVE-VQC & YouTube \\
%     \midrule
%     MAE         & 8.29 & 0.7169 & 0.7807 & 0.7631 \\
%     Swin-B      & 0.72 & 0.7892 & 0.7701 & 0.7816 \\
%     X3D         & 2.35 & 0.6165 & 0.5995 & 0.6432 \\
%     ir-CSN152   & 1.41 & 0.7647 & 0.6304 & 0.7349 \\
%     CLIP        & 2.49 & 0.8398 & 0.7832 & 0.8089 \\
%     ConvNeXt    & 0.62 & 0.7794 & 0.7554 & 0.7988 \\
%     TimeSformer & 4.47 & 0.8044 & 0.7427 & 0.7541 \\
%     ViT-B       & 3.15 & 0.7879 & 0.7353 & 0.7218 \\
%     \bottomrule
%     \end{tabular}
%     \label{tab:dbi1}
% \end{table}

\begin{table}[t]
    \centering
    \footnotesize
    \caption{Ablation on the effectiveness of DBI.}
    \vspace{-3mm}
    \begin{tabular}{c|c|cc}
    \toprule
    Datasets & $\omega_n$ & PLCC & SRCC \\
    \midrule
    KoNViD-1k & $1/N$    & 0.8631 & 0.8521 \\
              & $1/\psi$ & \textbf{0.8718} & \textbf{0.8568} \\
    \midrule
    LIVE-VQC  & $1/N$    & \textbf{0.8205} & 0.8107 \\
              & $1/\psi$ & 0.8198 & \textbf{0.8110} \\
    \midrule
    YouTube-UGC & $1/N$    & 0.8427 & 0.8446 \\
                & $1/\psi$ & \textbf{0.8570} & \textbf{0.8578} \\
    \bottomrule
    \end{tabular}
    \label{tab:dbi2}
    \vspace{-3mm}
\end{table}

\paragraph{Ablation on the DBI strategy.} The effectiveness of DBI can be evaluated in two aspects: (1) Model selection strategy. We performed 10 experiments based on randomly selected pretrained models in KoNViD-1k, resulting in PLCC ($0.7917 \pm 0.0578$, SRCC ($0.7583 \pm 0.0492$). Compared with the DBI-based strategy, the performance is poor and the randomness is high. (2) Feature integration. Tab.~\ref{tab:dbi2} shows the effectiveness of DBI in guiding the integration of different models, allowing more relevant models to contribute more. 

\section{Conclusion}

In this paper, we proposed PTM-VQA that utilizes in-the-wild pretrained models as feature extractors for NR-VQA tasks, transferring quality-related knowledge from diverse pre-text domains. The DBI scores are used to select candidates from a large amount of available pretrained models. To constrain features with large diversity into a unified latent space of quality and tackle outliers, we propose a new ICID loss. Under small computational cost, PTM-VQA models obtain SOTA results in widely-used benchmarks. Experiments in larger datasets and cross-database evaluation further prove generalizability.

% Furthermore, how to further use the pretrained models is still an open question with great practical significance. There are some problems worthy of research, which we would like to explore in future work: (1) is there a more effective way to select models? (2) We find that different datasets require different model combinations (by trial-and-error) for optimal performance. Is there an automatic selection manner? (3) What exactly did the pretrained models migrate? We hope to inspire subsequent related research.

\section*{Acknowledgments}

This work was partly supported by the Natural Science Foundation of China (NSFC) under Grant No. 62306309.

\clearpage

{
    \small
    \bibliographystyle{ieeenat_fullname}
    \bibliography{main}
}


\end{document}